\documentclass[conference]{IEEEtran}
\IEEEoverridecommandlockouts

\usepackage{cite}
\usepackage{amsmath,amssymb,amsfonts}
\usepackage{graphicx}
\usepackage{textcomp}
\usepackage{xcolor}
\usepackage{bm}

\usepackage[linesnumbered,ruled,vlined]{algorithm2e}

\usepackage[breaklinks,colorlinks,citecolor=red]{hyperref}

\usepackage{float}
\usepackage{subcaption}

\DeclareMathOperator*{\argmax}{arg\,max}
\DeclareMathOperator*{\argmin}{arg\,min}

\newcommand{\colorblue}[1]{\textcolor{black}{#1}}  

\def\BibTeX{{\rm B\kern-.05em{\sc i\kern-.025em b}\kern-.08em
    T\kern-.2em\lower.7ex\hbox{E}\kern-.125emX}}
\begin{document}

\title{Distribution-Aware Mobility-Assisted Decentralized Federated Learning\\
}

\author{Md Farhamdur Reza$^1$, Reza Jahani$^1$, Richeng Jin$^{2}$, and Huaiyu Dai$^{1}$ \\
$^1$NC State University \quad  $^2$Zhejiang University\\
\texttt{\{mreza2, rjahani, hdai\}@ncsu.edu} \quad \texttt{richengjin@zju.edu.cn} \\
}

\maketitle

\begin{abstract}
Decentralized federated learning (DFL) has attracted significant attention due to its scalability and independence from a central server. 
In practice, some participating clients can be mobile, yet the impact of user mobility on DFL performance remains largely unexplored, despite its potential to facilitate communication and model convergence.
In this work, we demonstrate that introducing a small fraction of mobile clients, even with random movement, can significantly improve the accuracy of DFL by facilitating information flow. To further enhance performance, we propose novel distribution-aware mobility patterns, where mobile clients strategically navigate the network, leveraging knowledge of data distributions and static client locations. The proposed moving strategies mitigate the impact of data heterogeneity and boost learning convergence. Extensive experiments validate the effectiveness of induced mobility in DFL and demonstrate the superiority of our proposed mobility patterns over random movement. 
\end{abstract}


\section{Introduction}
The rapid growth of machine learning applications across diverse domains has driven increasing demand for efficient and scalable training frameworks. 
Traditional centralized frameworks rely on clients sending raw data to a central server for model training.
However, this approach has significant drawbacks, including high network traffic and privacy concerns, as raw personal data is transmitted over the network~\cite{chiang2016fog}.

Federated Learning (FL)~\cite{mcmahan2017communication} enables model training without sharing raw data, with each client training a local model and sending only updates to a central server for aggregation. 
However, FL faces several challenges, particularly data heterogeneity, where client data is often not independently and identically distributed (non-IID) in real-world scenarios. This heterogeneity leads to performance degradation, where conventional aggregation methods fail to generalize well across clients~\cite{feng2021federated}. 
To address this issue, various strategies have been proposed. For instance, FedAF~\cite{wang2024aggregation} improves aggregation by adapting weights based on data distribution, outperforming standard approaches like FedAvg~\cite{mcmahan2017communication} and FedProx~\cite{li2020federated}. Similarly, FedDisco~\cite{ye2023feddisco} introduces a distribution-aware strategy to address class imbalance across clients.
Another challenge is communication overhead, which grows significantly with more clients, leading to scalability concerns~\cite{barbieri2022decentralized}.

Decentralized Federated Learning (DFL) extends FL by eliminating the need for a global server, utilizing peer-to-peer communication for model aggregation~\cite{beltran2023decentralized}. 
In DFL, clients exchange model updates directly with their neighbors, defined by their communication range. Although this decentralized setup improves scalability and eliminates the reliance on a central server, it introduces new challenges.

\begin{figure}[t]
    \centering
    \includegraphics[width=0.85\linewidth]{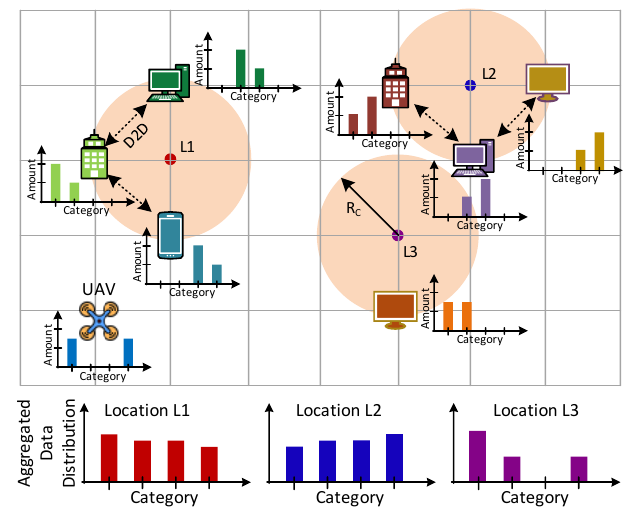} \vspace{-3mm}
    \caption{DFL schematic with a UAV as a mobile client and seven fixed clients. It illustrates the local data distributions of each client and the UAV’s movement across three locations (L1, L2, and L3). The shaded regions represent communication ranges, and the aggregated data distribution at each UAV location is shown at the bottom for reference.}
    \label{fig:sch_dia}  \vspace{-4mm}
\end{figure}

One critical factor affecting DFL performance is network topology~\cite{zhu2022topology, wang2019matcha}. 
\colorblue{Given a communication radius, a client participating in DFL is restricted to communicating with neighbors within the communication range. 
The structure of the resulting communication graph significantly influences information flow, model convergence, and overall performance.} 
In sparse network topologies, limited connectivity reduces the flow of updates, leading to slower convergence than denser topologies. Additionally, information may fail to propagate across disconnected networks, further hindering convergence.

\colorblue{In many real-world applications, clients are not always static; some exhibit mobility~\cite{yu2020mobility}, which influences communication patterns and information exchange in decentralized networks. Motivated by this, we investigate the impact of mobility in DFL to understand its effects on model performance and information flow.
Mobile clients move freely within the network following a predefined mobility pattern, while static clients remain fixed}. We first explore the impact of mobility by considering random movement and demonstrate that introducing mobile clients enhances performance by improving information dissemination across the network. These mobile clients explore the network, gather updates from fixed clients, and propagate information to other disconnected areas, mitigating the adverse effects of network sparsity.

To further utilize the benefits of mobility, we propose distribution-aware mobility patterns to enhance convergence by addressing both network sparsity and data heterogeneity. In this approach, mobile clients move based on knowledge of the data distributions and locations of fixed clients. Their trajectories are guided by an aggregated data distribution metric that determines effective movement paths to reduce heterogeneity and improve overall performance. 
A \textit{simple illustration} of the proposed mobility pattern is provided in Fig.~\ref{fig:sch_dia}, where an unmanned aerial vehicle (UAV) acts as a mobile client which is allowed to move across three locations: L1, L2, and L3. If L1 is the current location of the UAV, with the proposed approach, the UAV is most likely to move towards L3 to reduce the impact of data heterogeneity and sparsity. Our study further examines the impact of network connectivity parameters, such as communication radius, number of mobile clients, and their speed, on DFL performance.

To the best of our knowledge, this is one of the first works to investigate the impact of mobility in the DFL setting considering different networking parameters (such as communication radius, number of mobile clients, and velocity of mobile clients) and to design mobility patterns accordingly for performance improvement. Our main contributions are summarized as follows:
\begin{itemize}
    \item We demonstrate that mobility enhances performance in decentralized federated learning.
    \item We propose two distribution-aware mobility patterns, which outperform random movement and mitigate the issues caused by data heterogeneity.
    \item We conduct extensive experiments to reveal the impact of various networking parameters on overall model performance.
\end{itemize}

\section{Related Works}

Comprehensive convergence analyses for DFL setting are provided in~\cite{koloskova2020unified, jiang2025convergence}.
The positive correlation between the convergence rate and the spectral gap of the communication topology in decentralized settings is highlighted in \cite{zhu2022topology}.
To address the slow convergence rate, the authors in \cite{chen2021accelerating} propose Gossip SGD for distributed learning.
Moreover, a mutual knowledge transfer technique to mitigate performance degradation in DFL systems is introduced in \cite{li2021decentralized}. 
The authors in \cite{bellet2022d} demonstrate that a carefully designed topology can reduce the impact of data heterogeneity.
Meanwhile, a novel framework for mixing matrix optimization, aimed at minimizing gradient mixing error in decentralized settings, is proposed in \cite{dandi2022data}.
Similarly, the authors in \cite{le2023refined} consider the joint impact of topology and data heterogeneity on the convergence behavior of decentralized learning.

Existing work on the DFL setting has largely overlooked the impact of user mobility, despite its potential to significantly influence system performance. Several studies explore user mobility in the classic FL context~\cite{yu2020mobility, pervej2023resource, feng2022mobility, peng2023tame, bian2024accelerating}. 
Model aggregation strategies aimed at optimizing federated learning performance in vehicular networks are proposed in \cite{yu2020mobility, pervej2023resource}. 
The authors in \cite{feng2022mobility} investigate the impact of user mobility in classic FL and identify it as a negative factor causing performance degradation. In contrast, it is suggested in \cite{peng2023tame} that mobility can be beneficial when the users move at an appropriate speed.
The advantage of mobility in asynchronous FL is investigated in ~\cite{bian2024accelerating}.
However, none of these studies explore the effect of user mobility in fully decentralized settings.
Although existing work has explored the use of UAVs in the DFL setting, they often overlook the impact of mobility~\cite{qu2021decentralized}. In contrast, the author of \cite{de2024exploring} explores user mobility in DFL, but only in terms of random movements.


\section{System Model}
We consider a decentralized network of $C$ clients denoted by the set $\mathcal{C} = [C] := \{1,..., C\}$. 
Among these, a subset $\mathcal{C}_m \subset \mathcal{C}$ comprises mobile clients that move according to a predefined mobility pattern, while the remaining clients, denoted as $\mathcal{C}_s = \mathcal{C} \setminus \mathcal{C}_m$, remain static throughout the process.
Each client $i \in \mathcal{C}$ holds a local dataset $\mathcal{D}_i= \{(\bm s_n, y_n) \mid  n=1,2,...,|\mathcal{D}_i|\}$ and updates its local model $x_i$ based on its local dataset, where $\bm s_n$ and $y_n$ denote the $n$-th input and corresponding true label. 
DFL objective is to optimize the following function~\cite{koloskova2020unified}: \vspace{-2mm}
\begin{equation}
    f^* := \Big[\min_{x \in \mathbb{R}^d} f(x) := \frac{1}{|\mathcal{C}|} \sum_{i \in \mathcal{C}} f_i(x) \Big],  \vspace{-2mm}
\end{equation}
where $f_i(x) := \mathbb{E}_{\xi_i \sim \mathcal{D}_i} F_i(x; \xi_i)$ represents the local expected loss over mini-batches $\xi_i$, with $F_i(.)$ denoting the client-specific loss function, and $d$ being the model dimensionality. 

Each client $i$ has a limited communication range defined by a circular coverage area of radius $R_c$. 
A client $i$ can communicate with another client $j \in \mathcal{C}\setminus \{i\}$ if $j$ lies within this coverage.
The set of connected neighbors for client $i$ at global round $t$ is given by: $\mathcal{N}_i^{(t)} = \{j \mid \| L_i^{(t)} - L_j^{(t)}\|_2 \leq R_c, \forall j \in \mathcal{C}\setminus\{i\}\}$, where $L_i^{(t)}$ represents the location of client $i$ at the global round $t$. 
While fixed clients remain at the same locations, i.e., $L_i^{(t-1)} = L_i^{(t)}$, $\forall i \in \mathcal{C}_s$, mobile clients change their locations according to a mobility pattern constrained by the maximum movable radius $R_m$. Consequently, the network topology varies over time due to mobile clients. 

In this decentralized framework, each client maintains local parameters $x^{(t-1)}_i \in \mathbb{R}^d$ and computes the local stochastic gradient $g^{(t-1)}_i:= \nabla F_i(x^{(t-1)}_i, \xi_i)$ based on samples from its dataset at round $t$. 
Following this, each client $i \in \mathcal{C}$ exchanges models with its neighbors $\mathcal{N}^{(t)}_i$ through D2D communication and aggregates the received models. The decentralized training process involves two phases per round:
\begin{enumerate}
    \item \textit{Local update}: $x_i^{(t-\frac{1}{2})} = x_i^{(t-1)} - \eta g_i^{(t-1)}$, where $\eta$ is the learning rate.
    \item \textit{Consensus update}: $x_i^{(t)} = \sum_{j=1}^N w^{(t)}_{i,j} x_j^{t-\frac{1}{2}}$, where $w^{(t)}_{i,j}$ is the $(i,j)$-th entry of the mixing matrix $\bm W^{(t)} \in \mathbb{R}^{C \times C}$, satisfying $w^{(t)}_{i,j}=0$ if $j \notin \mathcal{N}^{(t)}_i \cup \{i\}$.
\end{enumerate}
For convenience, we use the following matrix notation to stack all the models, 
$
    \bm X^{(t)} = \big[ x_1^{(t)},..., x_C^{(t)} \big] \in \mathbb{R}^{d \times C}.
$
Likewise, we define $\bm {G}^{(t)} = [g_1^{(t)},..., g_C^{(t)}]$. Hence, the local and consensus updates for all the models can be expressed as: $\bm X^{(t-\frac{1}{2})} = \bm X^{(t-1)} - \eta \bm G^{(t-1)}$ and $\bm X^{t} = \bm X^{(t-\frac{1}{2})} \bm W^{(t)} $, respectively, where $\bm W^{(t)}$ is the time-varying mixing matrix due to the mobile clients. 
Non-zero weights of the consensus matrix $\bm W^{(t)}$ are determined based on the local dataset sizes of neighboring clients, following FedAvg~\cite{mcmahan2017communication}. The $(i,j)$-th entry of $\bm W^{(t)}$ is as follows: \vspace{-2mm}
\begin{equation}
w_{i,j}^{(t)} = 
    \begin{cases}
        \frac{|\mathcal{D}_j|}{|\mathcal{D}_i|+\sum_{k \in \mathcal{N}_i^{(t)}} |\mathcal{D}_k|} , & \text{if  }\; j \in \mathcal{N}_i^{(t)} \cup \{i\},\\
        0 , & \text{otherwise}.
    \end{cases}
\end{equation}


\section{Mobility Models}

In the mobility model, we consider clients positioned on a $G \times G$ grid, where each location is represented as $ (p,q) \in \mathcal{G_L}$ with $\mathcal{G_L} = \{(p,q) \mid p,q \in \{1,2,...,G\}\}$. 
At the beginning of the training process, each user $i$ is randomly assigned an initial location $L_i^{(0)} \in \mathcal{G_L}$ within the grid.
Based on the mobility pattern, each mobile client in $\mathcal{C}_m$ relocates to a new grid position in each global round, and all clients in $\mathcal{C}$ exchange their models with their neighboring clients for aggregation. 
In this section, we discuss baseline mobility models and elaborate on the proposed distribution-aware mobility patterns.

\subsection{Baseline Mobility Patterns}
We consider two baseline models: \textit{Static} and \textit{Random Movement}. 
For the \textbf{Static} scenario, client locations remain fixed throughout the training, i.e., $L_i^{(t-1)} = L_i^{(t)}; \forall i \in \mathcal{C}, \forall t \in [T]$, where $T$ is the total number of global training rounds. 
For the \textbf{Random Movement}, mobile clients $i_m \in \mathcal{C}_m$ uniformly randomly select a new location $L_{i_m}^{(t)}$ within the grid at each round $t$, subject to their location at the end of round $t-1$ and a maximum allowable displacement per round $R_m$.
This constraint ensures that clients move within a feasible range per round, maintaining realistic mobility patterns. Formally,
\begin{equation}
    L_{i_m}^{(t)} \sim \text{Uniform}\Big(\{L \mid \|L-L_{i_m}^{(t-1)}\|_2 \leq R_m, L \in \mathcal{G_L} \}\Big).
\end{equation}

\subsection{Proposed Mobility Patterns} \vspace{-1mm}
Before illustrating our proposed mobility patterns, we first introduce the term \textit{distribution distance}, which integrates heterogeneity and mobility. 
In FL, it is commonly assumed that the central server has knowledge of mobile clients' speeds and locations~\cite{yu2020mobility}. Additionally, FedDisco~\cite{ye2023feddisco} assumes that the server is aware of each client's category-wise data distribution.
Building on this, we assume that mobile clients in $\mathcal{C}_m$ have knowledge of the category-wise sample distributions and locations of static clients in $\mathcal{C}_s$. While these are strong assumptions, we will relax them in future work. 
These assumptions allow mobile clients to calculate the data distribution across different grid points in the network. 
Given the category-wise distribution of static clients and their locations, a mobile client $i_m \in \mathcal{C}_m$ can determine the data distribution at any location in $\mathcal{G_L}$ before starting training. 
The category-wise distribution at a location $L \in \mathcal{G_L}$, if the mobile client moves there, is given by: \vspace{-2mm}
\begin{equation} 
    D_{i_m,l}^{L} = \frac{\sum_{c \in {\mathcal{\zeta}}_{i_m,L}} | \{ (\bm s_n, y_n) \in \mathcal{D}_c \mid y_n = l \}|}{\sum_{c \in {\mathcal{\zeta}}_{i_m,L}} |\mathcal{D}_c|}, l \in [Y],
    \label{eq:cat_dis}
\end{equation}
where ${\mathcal{\zeta}}_{i_m,L} := \{j \mid \|L_j^{(0)} - L \|_2 \leq R_c, j \in \mathcal{C}_s\} \cup \{i_m\}$ denotes the set of fixed clients within the coverage area of the mobile client $i_m$ at location $L$, including $i_m$ itself, and $[Y]$ denotes the set of classification labels. 

Using this category-wise distribution, we define the \textit{distribution distance} from the client's current location $L_{i_m}^{(t-1)}$ (considered as the reference point) at the start of round $t$ as:
\begin{equation}
    d_{L_{i_m}^{(t-1)}}^{L} = \sqrt{\sum_{l \in [Y]} \left(  D_{i_m,l}^{L}- D_{i_m,l}^{L_{i_m}^{(t-1)}} \right)^2 }, \quad \forall L \in \mathcal{G_L}.
    \label{eq:distr_dis}
\end{equation}

\subsubsection{Distribution-Aware Mobility (DAM)}
Given the distribution distances across all grid points with respect to the current location $L_{i_m}^{(t-1)}$ of mobile client $i_m$, our proposed \textbf{Distribution-Aware Mobility (DAM)} \colorblue{pattern assigns higher probabilities to locations with greater distribution distances. This encourages movement toward regions that mitigate the impact of data heterogeneity. 
Based on this probability distribution, the next desired location is determined. 
The probability of mobile client $i_m$ moving to a location $L \in \mathcal{G_L}$ at the end of local update of round $t$ is given by:}
\begin{equation} 
p^{(t)}_{ L | L_{i_m}^{(t-1)}} = \frac{d_{L_{i_m}^{(t-1)}}^{L}}{\sum_{L \in \mathcal{G_L}}d_{L_{i_m}^{(t-1)}}^{L}}, \quad \forall L \in \mathcal{G_L} . 
\label{eq:dam_prob_distr}
\end{equation}

Therefore, the desired new location of $i_m$ can be sampled as: $L_{i_m}^{(des., t)}\sim p^{(t)}_{ L | L_{i_m}^{(t-1)}}$. However, the movement of mobile clients is restricted by a maximum allowable displacement per round $R_m$. 
If $ \|L_{i_m}^{(des., t)}  - L_{i_m}^{(t-1)}\|_2 > R_m$, the client moves to the nearest location to $L_{i_m}^{(des., t)}$ along its path. Formally, the updated location at the end of round $t$'s local update is:
\begin{equation}
     L_{i_m}^{(t)} = \begin{cases}
         L_{i_m}^{(des., t)}; \quad \text{if } \|L_{i_m}^{(des., t)}  - L_{i_m}^{(t-1)}\|_2 \leq R_m\\
        \argmin_{L \in \Gamma} \|L-L_{i_m}^{(des., t)} \|_2; \quad \text{otherwise},
     \end{cases}
     \label{eq:obtained_loc}
\end{equation}
where $\Gamma := \{ L \in \mathcal{G_L} \mid \|L-L_{i_m}^{(t-1)}\|_2 \leq R_m\}$. Nevertheless, if the mobile client $i_m$ does not reach its destination within the round, its destination remains unchanged until it reaches it. 
Once the desired location is reached, a new destination is drawn according the distribution in \eqref{eq:dam_prob_distr}.
Thus, the desired location can be expressed as:
\begin{equation}
     L_{i_m}^{(des., t)} \begin{cases}
         \sim p^{(t)}_{ L | L_{i_m}^{(t-1)}}; ~ &\text{if } L_{i_m}^{(t-1)} = L_{i_m}^{(des., t-1)} \\
        = L_{i_m}^{(des., t-1)}; ~ &\text{otherwise}.
     \end{cases}
     \label{eq:desired_loc}
\end{equation}

\subsubsection{Distribution-Aware Cluster-Center Mobility (DCM)}
The destination determined by the DAM mobility pattern—based on the probability distribution in Eq.~\eqref{eq:dam_prob_distr}—can be any location within the grid, i.e., $L_{i_m}^{(des., t)} \in \mathcal{G_L}$.
However, the large search space can potentially slow the convergence in mobility-assisted DFL.
To address this, we propose \textbf{Distribution-Aware Cluster-Center Mobility (DCM)} pattern, which reduces the destination search space and further enhances convergence.
In DCM, mobile clients select destinations that correspond to the centers of clusters of static clients. These clusters are formed based on the communication radius $R_c$, and the cluster centers act as representative destinations.
The selection process follows a structured approach:
\begin{itemize}
    \item {\textit{Selecting the First Cluster Center}:}
    The first cluster center $L_{c_1}$ is sampled as a grid point $L_{c_1} \in \mathcal{G_L}$ where a mobile client $i_m \in \mathcal{C}_s$ can communicate with the maximum number of neighboring static clients within radius $R_c$: \vspace{-2mm}
    \[
      L_{c_1} \sim \argmax_{L \in \mathcal{G_L}}|\{j \in \mathcal{C}_s \mid \| L - L_j^{(0)} \|_2 \leq R_c\}|. \vspace{-2mm}
    \]
    \item \textit{Selecting Subsequent Cluster Centers}:
    The second cluster center is chosen from the remaining static clients (i.e., excluding those already covered by the first cluster).
    If multiple locations have the same maximum number of neighbors, the one covering the most neighbors among all static clients is selected. In case of a tie, a location is randomly chosen.
    This process repeats until all static clients are assigned to a cluster.
\end{itemize}
Algorithm~\ref{alg:clustering} formally describes this clustering procedure, ensuring an efficient selection of cluster centers $\mathcal{L}^c$.

\begin{algorithm}	[h] \small
	 \textbf{Inputs:} Communication Radius $R_c$; Grid locations $\mathcal{G_L}$; Static clients $\mathcal{C}_s$; Clients initial locations  $L_i^{(0)} \in \mathcal{G_L}, \forall i \in \mathcal{C}_s$.
	 
	 \textbf{Output:} Cluster Center Locations $\mathcal{L}^c$.
     
	 $\mathcal{L}^c \gets \{\}$; \quad $\mathcal{C}_s^r \gets \mathcal{C}_s$; \quad $n=1$\\

    \While{$|\mathcal{C}^r_s| > 0 $}{
        $\mathcal{L}_d^c = \argmax_{L \in \mathcal{G_L}}|\{j \in \mathcal{C}^r_s \mid \| L - L_j^{(0)} \|_2 \leq R_c\}|$ \\
        \If{$|\mathcal{L}_d^c| > 1 ~\&~ n>1$}{
            $\mathcal{L}_d^c  = \argmax_{L \in \mathcal{L}_d^c}|\{j \in \mathcal{C}_s \mid \| L - L_j^{(0)} \|_2 \leq R_c\}|$ \\
        }
        $L_{c_n} \sim \mathcal{L}_d^c$  \\
        $\mathcal{C}^r_s = \mathcal{C}^r_s \setminus \{j \in \mathcal{C}_s \mid \| L_{c_n} - L_j^{(0)} \|_2 \leq R_c\}$\\
        $\mathcal{L}^c = \mathcal{L}^c \cup \{L_{c_n}\}$; \quad $n= n+1$
    }
    \caption{Clustering}
    \label{alg:clustering}
\end{algorithm}

In DCM, the mobile clients are destined to move across cluster centers $\mathcal{L}^c$, which reduces the number of possible desired locations compared to DAM. 
The distribution distance $d_{L_{i_m}^{(t-1)}}^{L^c}$ of each cluster center $L^c \in \mathcal{L}^c$ with respect to a mobile client $i_m$'s current location $L_{i_m}^{(t-1)}$ at round $t$ is computed using the same expression in Eq.~\eqref{eq:distr_dis}.
Thus, the probability distribution for movement across cluster centers, based on the distribution distance, is: \vspace{-2mm}
\begin{equation} 
p^{(t)}_{ L^c | L_{i_m}^{(t-1)}} = \frac{d_{L_{i_m}^{(t-1)}}^{L^c}}{\sum_{L^c \in \mathcal{L}^c}d_{L_{i_m}^{(t-1)}}^{L^c}}, \quad \forall L^c \in \mathcal{L}^c . 
\label{eq:dcm_prob_distr} \vspace{-2mm}
\end{equation}
Due to the mobility constraint $R_m$, the movement of $i_m \in \mathcal{C}_m$ across $\mathcal{L}^c$ exploiting the probability distribution in Eq.~\eqref{eq:dcm_prob_distr} follows similar expressions as in Eq.~\eqref{eq:desired_loc} and Eq.~\eqref{eq:obtained_loc} to obtain the desired and the updated locations at round $t$, respectively.

All the mobility patterns discussed are constrained by the movement constraint $R_m$. In the Random Movement pattern, mobile clients select random locations within the mobility constraint $R_m$. In contrast, with the proposed DAM and DCM patterns, mobile clients move toward their desired locations $L_{i_m}^{(des., t)}$, and have to choose an intermediate location along the route if $\|L_{i_m}^{(des., t)}  - L_{i_m}^{(t-1)}\|_2 > R_m$ while performing local updates in round $t$. As a extreme case, these mobility patterns can be considered as \textbf{unconstrained movement} when $R_m=\infty$, allowing mobile clients to move freely to any desired location within the network in a given round. 

\begin{figure*}[t] \vspace{0mm}
    \begin{subfigure}{0.24\textwidth}
        \includegraphics[width=0.9\textwidth]{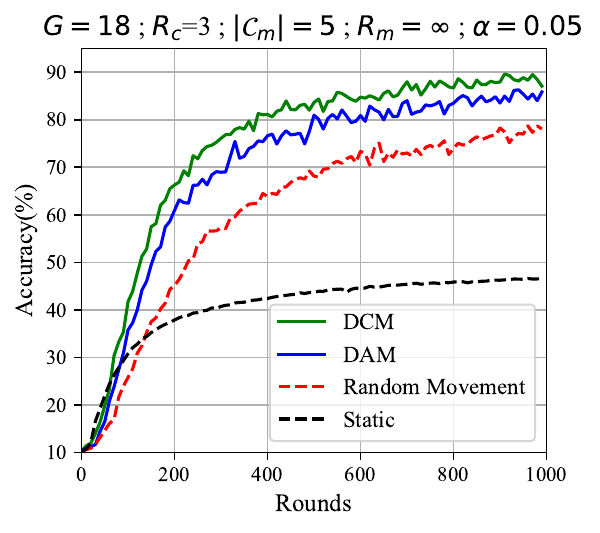}
        \caption{Unconstrained movements}
        \label{fig:acc_uncons_mnist_alpha0.05}
    \end{subfigure}
    \begin{subfigure}{0.24\textwidth}
        \includegraphics[width=0.9\textwidth]{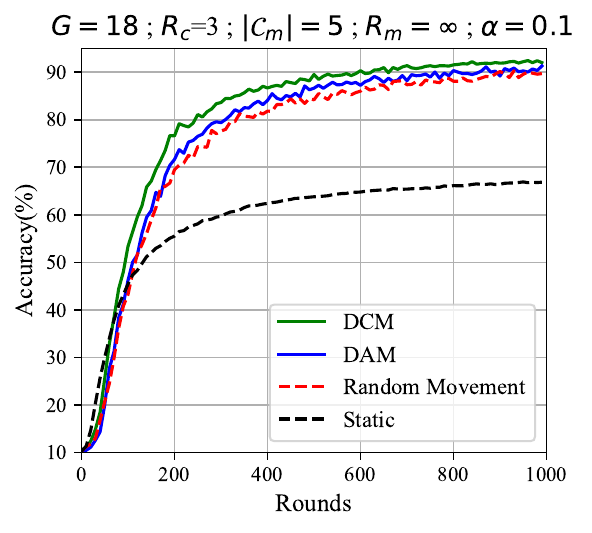}
        \caption{Unconstrained movements}
        \label{fig:acc_uncons_mnist_alpha0.01}
    \end{subfigure}
    \begin{subfigure}{0.24\textwidth}
        \includegraphics[width=0.9\textwidth]{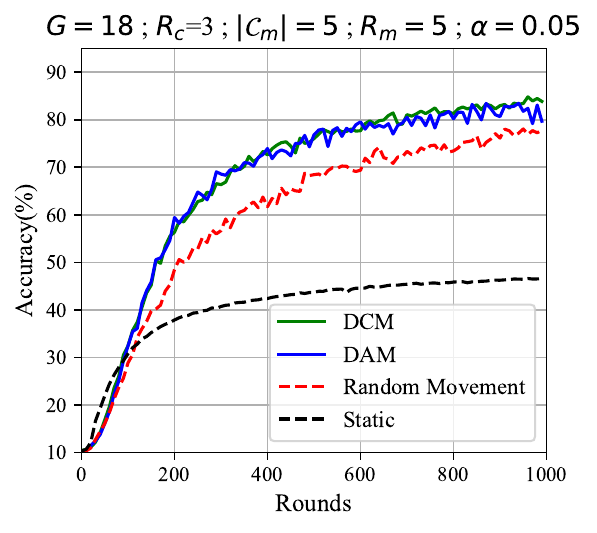}
        \caption{Constrained movements}
        \label{fig:acc_cons_mnist_alpha0.05}
    \end{subfigure}
    \begin{subfigure}{0.24\textwidth}
        \includegraphics[width=0.9\textwidth]{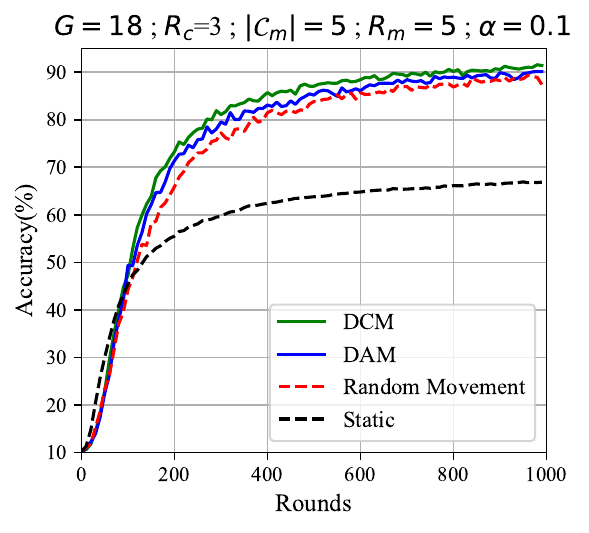}
        \caption{Constrained movements}
        \label{fig:acc_cons_mnist_alpha0.01}
    \end{subfigure}  \vspace{-2mm}
    \caption{Accuracy of DFL system on MNIST Dataset after 1000 rounds with 20 clients.} \vspace{-4mm}
    \label{fig:acc}
\end{figure*}

\begin{figure*}
    \begin{subfigure}{0.32\textwidth}
        \centering
        \includegraphics[width=0.7\linewidth]{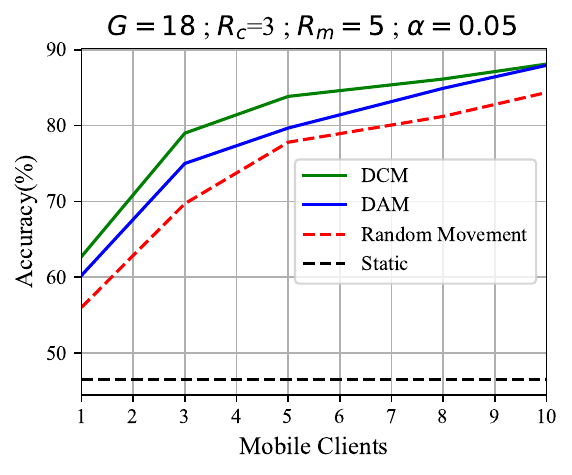}
        \caption{Accuracy vs. $|\mathcal{C}_m|$}
        \label{fig:mc}
    \end{subfigure} \hfill
    \begin{subfigure}{0.32\textwidth}
        \centering
        \includegraphics[width=0.7\linewidth]{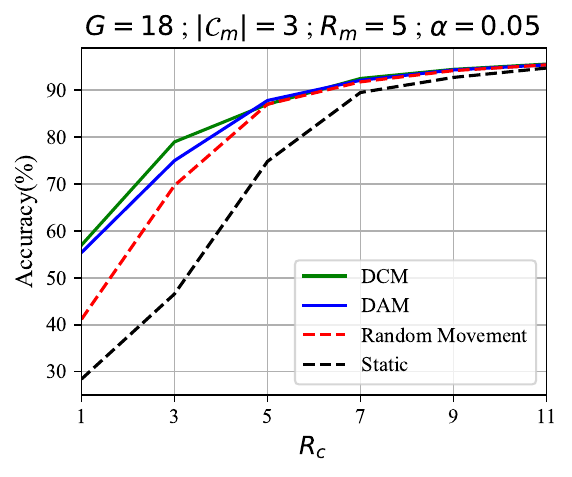}
        \caption{Accuracy vs. $R_c$}
        \label{fig:rc}
    \end{subfigure}
    \begin{subfigure}{0.32\textwidth}
        \centering
        \includegraphics[width=0.7\textwidth]{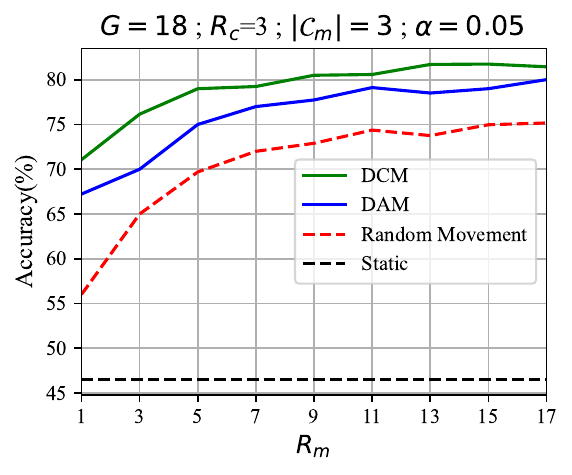}
        \caption{Accuracy vs. $R_m$}
        \label{fig:rm}
    \end{subfigure} \vspace{-2mm}
    \caption{Impact of $|\mathcal{C}_m|$, $R_c$ and $R_m$ on accuracy of DFL system on MNIST Dataset after 1000 rounds with 20 clients.} \vspace{-4mm}
\end{figure*}




\section{Experimental Results}
In this section, we conduct experimental evaluations to assess our proposed mobility patterns in a DFL system and compare their performance against \textbf{Static} and \textbf{Random Movement}, which serve as baselines.

\subsection{Simulation Setting}
We divide our experiments into three different settings. In the first setting, we consider the MNIST classification task on a network of size $18\times18$ with 20 clients. 
In the second setting, we scale our simulations to a network of size $35\times35$ with 50 clients, again using the MNIST dataset, to evaluate the scalability of our framework. 
Finally, in the third setting,  we assess the generalizability of our approaches by training on the CIFAR-10 dataset with 20 clients and a network of size $18 \times 18$.
Since data heterogeneity is a crucial factor in DFL, we distribute the training samples among clients following a Dirichlet distribution with parameter $\alpha$ for local updates and use the corresponding test sets for evaluation. Unless otherwise specified, we consider communication radius $R_c=3$ for the network topology. 

For training on the MNIST dataset, we consider a simple model structure across the clients: two convolution layers followed by two fully connected layers. Training is performed using full-batch gradient descent with a learning rate of 0.3. 
For the CIFAR-10 dataset, in contrast, we adopt a ResNet50 model pretrained on ImageNet-1K, adapting the first convolutional layer (kernel size 3, stride 1) for CIFAR-10 input, and removing the initial MaxPool layer.
Training on CIFAR-10 is conducted using mini-batch gradient descent with a batch size of 512 and a learning rate of 0.01.
For both datasets, we use the SGD optimizer with momentum and weight decay values of 0.9 and 0.0005, respectively.
To evaluate performance, we compute the average test accuracy across clients over 10 Monte Carlo runs, where models are trained on the respective training datasets (MNIST and CIFAR-10) and evaluated on the corresponding test sets every 10 global rounds.

\subsection{Results Analysis} \vspace{-1mm}
Fig.~\ref{fig:acc} shows the DFL system's accuracy after 1,000 training rounds on MNIST dataset in a network of size 18×18, with 20 clients, including $|\mathcal{C}_m|=5$ mobile clients. 
Fig.~\ref{fig:acc_uncons_mnist_alpha0.05} and Fig.~\ref{fig:acc_cons_mnist_alpha0.05} show the results for $\alpha=0.05$, while Fig. ~\ref{fig:acc_uncons_mnist_alpha0.01} and Fig.~\ref{fig:acc_cons_mnist_alpha0.01} correspond to $\alpha$ = 0.1, representing a less heterogeneous scenario. Additionally, we compare the performance for both unconstrained ($R_m=\infty$) and constrained ($R_m=5$) mobility.


The results show that introducing mobility into the network improves performance. Specifically, when a subset of clients are mobile and follow the Random Movement, the model achieves significantly higher accuracy compared to the Static scenario, where all users remain fixed at initial locations. Furthermore, considering Random Movement as a baseline, DAM and DCM improve accuracy on average by approximately 8\% in both constrained and unconstrained movements, in highly heterogeneous settings ($\alpha = 0.05$). Similarly, when the data is distributed with less heterogeneity ($\alpha = 0.1$), DAM and DCM still achieve 3\% accuracy improvement over the baseline. As expected, the proposed DCM outperforms DAM, and unconstrained mobility patterns lead to better convergence compared to their constrained counterparts.

\subsubsection{Impact of Mobile Clients}
To investigate further on the effect of the number of mobile clients $|\mathcal{C}_m|$ in the network, we consider the same learning task on the MNIST dataset in a network with 20 clients, data heterogeneity parameter $\alpha = 0.05$, and mobility constraint $R_m=5$. 
As shown in Fig.~\ref{fig:mc}, increasing the number of mobile clients exposes more fixed clients to new updates, which enhances overall model performance. Additionally, both DCM and DAM consistently outperform Random Movement across different $|\mathcal{C}_m|$ values. 
As can be seen, DCM and DAM achieve, on average, 7\% and 5.3\% better accuracy, respectively, compared to Random Movement across diferent values of $|\mathcal{C}_m|$.

\subsubsection{Impact of Communication Radius ($R_c$)}
Fig. ~\ref{fig:rc} illustrates the impact of communication range in a network with 20 clients, where $|\mathcal{C}_m|=3$ mobile clients are deployed in a network, and data heterogenity parameter $\alpha=0.05$, and the mobility constraint is set to $R_m=5$. 
As expected, with a larger communication range, more clients are allowed to exchange their model updates. Therefore, higher accuracy is obtained. However, it can be seen that if the mobile clients are utilized with our proposed mobility patterns, they can outperform Random Movement significantly in networks with low connectivity. For example, at $R_c=1$, DCM and DAM achieve around 15\% higher accuracy than Random Movement.

\subsubsection{Impact of Mobility Constraint ($R_m$)}
To investigate the effect of client mobility constraint $R_m$, we conduct the simulation in a network with 20 clients and 3 mobile clients with $\alpha = 0.05$ for data heterogeneity and repeat these experiments while utilizing the mobile clients at different speeds. 
Fig.~\ref{fig:rm} illustrates the impact of $R_m$ and shows that accuracy improves with the increase of $R_m$ and gradually approaches the performance of mobile clients operating without movement constraints in the same setting. This is because mobile clients with higher $R_m$ are allowed to explore the network with more freedom and provide the fixed clients with more diverse and useful updates.

\begin{figure}[t] \vspace{-2mm}
    \begin{subfigure}{0.24\textwidth}
        \includegraphics[width=0.9\linewidth]{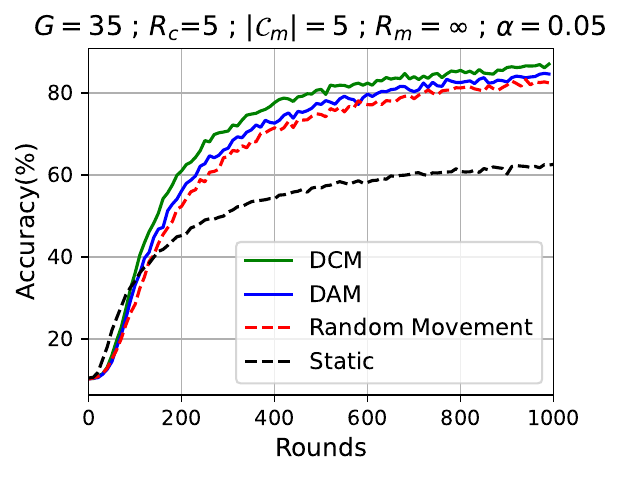}  
        \caption{Unconstrained movements}
    \end{subfigure}
    \hfill
    \begin{subfigure}{0.24\textwidth}
        \includegraphics[width=0.9\linewidth]{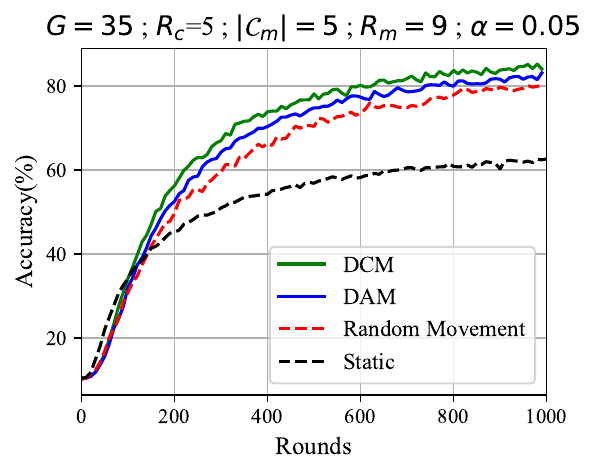}
        \caption{Constrained movements}
    \end{subfigure}
    \caption{Accuracy of DFL with 50 clients after 1000 rounds in a larger network with 5 mobile clients using MNIST dataset.}
    \label{fig:tc50} \vspace{-4mm}
\end{figure}

\subsubsection{Larger Network}
To evaluate the scalability of our proposed mobility patterns, we simulate the DFL system in a larger network with a size of 35$\times$35 and 50 clients by utilizing 5 mobile clients and mobility constraint $R_m=9$. We set $R_c=5$ and considered $\alpha = 0.05$ to investigate our proposed framework in a highly heterogeneous situation. Fig.~\ref{fig:tc50} illustrates the accuracy of DFL in 1,000 rounds with this setting. The proposed DAM and DCM mobility patterns outperform Random Movement in both unconstrained and constrained settings. Specifically, DCM achieves approximately 5\% and 4\% higher accuracy than Random Movement in the uncostrained and constrained scenarios, respectively.

\begin{figure}[t]
    \begin{subfigure}{0.24\textwidth}
        \centering
        \includegraphics[width=0.9\linewidth]{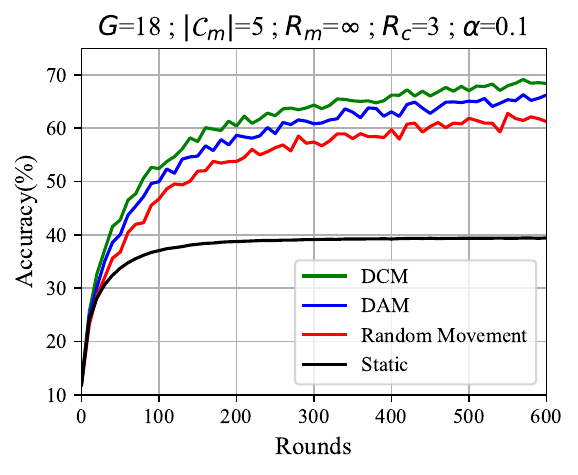}
        \caption{Unconstrained movements}
    \end{subfigure}
    \hfill
    \begin{subfigure}{0.24\textwidth}
        \centering
        \includegraphics[width=0.9\linewidth]{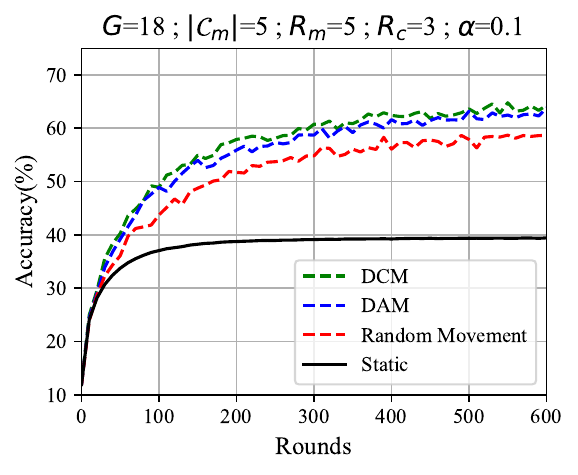}
        \caption{Constrained movements}
    \end{subfigure} \vspace{-4mm}
    \caption{Accuracy of DFL with 20 clients after 600 rounds in a network with 5 mobile clients using CIFAR-10 dataset.}
    \label{fig:cifar} \vspace{-4mm}
\end{figure}

\subsubsection{Performance on CIFAR-10}
We conduct additional experiments on the CIFAR-10 dataset using a $18 \times 18$ network with 20 clients, $R_c = 3$, and $\alpha = 0.1$. We compare the performance of the proposed DAM and DCM under both constrained ($R_m=5$) and unconstrained movement scenarios. As in Fig.~\ref{fig:cifar}, mobility significantly enhances convergence for CIFAR-10, mirroring the improvements observed on MNIST. 
Furthermore, in the unconstrained scenario, DAM and DCM achieve approximately 5\% and 8\% higher accuracy than random movement, while in the constrained setting, they offer around 4\% and 5\% higher accuracy, respectively.

\section{Conclusion and Future Works} \vspace{-1mm}
In this work, we explored the impact of mobility in DFL systems and demonstrated that the presence of mobile clients, even with random movement, significantly improves accuracy. To this end, we proposed two mobility patterns, DAM and DCM, which allow the mobile client to explore the network more effectively, leveraging the data distribution and location of static clients, than the baselines. 
Experimental results demonstrate that in highly heterogeneous data settings, both DAM and DCM outperform random movement, with DCM achieving particularly notable accuracy improvements.
\colorblue{For future work, we plan to theoretically analyze convergence and relax the assumption that mobile clients know static clients' data distributions and locations, exploring mobility’s impact.}


\bibliographystyle{IEEEtran}  
\bibliography{main.bib}  

\vspace{12pt}

\end{document}